\documentclass[final,1p,times]{elsarticle}

\usepackage{hyperref}       
\usepackage{url}            
\usepackage{booktabs}       
\usepackage{nicefrac}       
\usepackage{microtype}      
\usepackage{xcolor}         

\usepackage{amsfonts,amsmath,amsthm,amssymb}
\usepackage{bm}
\usepackage{braket}
\usepackage{float}
\usepackage{graphicx}
\usepackage{makecell}
\usepackage{multirow}
\usepackage{subcaption}

\newtheorem{theorem}{Theorem}

\journal{Pattern Recognition}

\begin{document}

\begin{frontmatter}

\title{Negational Symmetry of Quantum Neural Networks for Binary Pattern Classification}

\author[1]{Nanqing Dong\corref{cor1}}
\cortext[cor1]{Corresponding author}
\ead{nanqing.dong@cs.ox.ac.uk}
\author[2]{Michael Kampffmeyer}
\author[1]{Irina Voiculescu}
\author[3,4]{Eric Xing}

\affiliation[1]{organization={Department of Computer Science, University of Oxford},
            city={Oxford},
            postcode={OX1 3QD}, 
            country={UK}}
\affiliation[2]{organization={Department of Physics and Technology, UiT The Arctic University of Norway},
            city={Troms{\o}},
            postcode={9019}, 
            country={Norway}}
\affiliation[3]{organization={Machine Learning Department, Carnegie Mellon University},
            city={Pittsburgh},
            postcode={PA 15213}, 
            country={USA}}
\affiliation[4]{organization={Mohamed bin Zayed University of Artificial Intelligence},
            city={Masdar City, Abu Dhabi},
            country={UAE}}

\begin{abstract}
Although quantum neural networks (QNNs) have shown promising results in solving simple machine learning tasks recently, the behavior of QNNs in binary pattern classification is still underexplored. In this work, we find that QNNs have an Achilles' heel in binary pattern classification. To illustrate this point, we provide a theoretical insight into the properties of QNNs by presenting and analyzing a new form of \textit{symmetry} embedded in a family of QNNs with \textit{full entanglement}, which we term \textit{negational symmetry}. Due to negational symmetry, QNNs can not differentiate between a quantum binary signal and its negational counterpart. We empirically evaluate the negational symmetry of QNNs in binary pattern classification tasks using Google's quantum computing framework. Both theoretical and experimental results suggest that negational symmetry is a fundamental property of QNNs, which is not shared by classical models. Our findings also imply that negational symmetry is a double-edged sword in practical quantum applications.
\end{abstract}

\begin{keyword}
Deep learning \sep Quantum machine learning \sep Binary pattern classification \sep Representation learning \sep Symmetry
\end{keyword}

\end{frontmatter}


\section{Introduction}
\label{sec:intro}
In contrast to \emph{quantum-inspired machine learning}, which incorporates the concepts of quantum mechanics with classical machine learning (ML)~\cite{bai2015quantum,li2017quantum,zhang2019quantum}, quantum machine learning (QML)~\cite{schuld2015introduction,biamonte2017quantum} aims to understand how to devise and implement ML algorithms on quantum computers.
QML has received increasing attention due to the quantum supremacy experiment \cite{arute2019quantum} on near-term noisy intermediate-scale quantum (NISQ) \cite{preskill2018quantum} devices. However, the study of QML is still in an early stage. While various classical ML methods, such deep neural networks (DNNs) and convolutional neural networks (CNNs)~\cite{gu2018recent}, can easily show robust performance in binary pattern classification~\cite{watanabe1993relationships,badu2007classification,langseth2009latent}, a fundamental task that is used to evaluate the efficiency of ML algorithms~\cite{santhanam2016non}, quantum neural networks (QNNs)~\cite{mcclean2018barren,kerenidis2020quantum,bausch2020recurrent} still have veiled mysteries. This paper aims to provide an exploratory understanding of the application of QNNs in binary pattern classification.

As a quantum circuit model consisting of \textit{unitary} quantum gates, QNNs have direct links with \textit{feed-forward neural networks} and \textit{invertible neural networks} \cite{jacobsen2018revnet} in classical ML. At present, the study of QNNs is still in an early phase as QNNs have not shown quantum supremacy in solving classical ML tasks. However, QNNs can be easily integrated with quantum devices and quantum data, which gives them unparalleled advantages in quantum applications without access to complex classical computing systems. This is often ignored when comparing QNNs with classical models. Moreover, although it appears that we can easily manipulate the conversion between classical and quantum data, the situation in a real quantum device is much more complicated than the simulated study. For example, the \textit{No-Cloning Theorem} \cite{wootters1982single} states that quantum data cannot be copied, which means that, for certain QML tasks, the quantum data cannot be converted for classical models. Thus, researchers are eager to peep into the blackbox of QNNs and to analyze the properties of these models.

In this work, we find that QNNs might not be able to differentiate between a binary pattern and its \emph{negational} counterpart, where the term \textit{negational} refers to the negation in bitwise operation or logical operation and the negational transformation of a binary pattern is equivalent to applying a \texttt{NOT} gate (negation operation) to all bits, i.e.~flip all bits. To illustrate this Achilles' heel of QNNs in binary pattern classification, we provide insights into a new form of symmetry inherent in a specific family of QNNs with \textit{full entanglement}\footnote{See Sec.~\ref{sec:prob} for the formal definition.}, i.e.~each data qubit is entangled with the readout qubit. We find that QNNs with full entanglement\footnote{For simplicity, we use QNNs and QNNs with full entanglement interchangeably in the rest of the paper, as it can be inferred from the context.} feature two kinds of symmetries for binary classification and the auxiliary representation learning tasks with \emph{quantum binary signals} as the input. A quantum binary signal (e.g.~a binary pattern) contains only $\ket{0}$ and $\ket{1}$, which is analogous to $0$ and $1$ (or black and white) in a classical setting. Here, we present two major findings. First, given a (quantum) binary pattern, a QNN with $Z$-measurement makes the same prediction for a binary pattern and its negational counterpart. Second, with the proposed quantum representation learning, we find that the learned feature vectors of a binary pattern and its negational counterpart are two opposite vectors. Essentially, QNNs are mathematical functions. Inspired by classical functional analysis\footnote{In a classical 2D Cartesian system, given a function $f$ and a variable $x$, if we have $f(x) = f(-x)$, then we say $f$ is symmetric to the line $x = 0$ or $f$ has reflectional symmetry. If we have $f(x) = -f(-x)$, then we say $f$ is symmetric to the origin point $(0, 0)$ or $f$ has rotational symmetry. The same logic applies in high-dimensional systems with function $f: \mathbb{R}^N \mapsto \mathbb{R}^M$ and $N$-dimensional vector $\bm{x}$.}, we denote this new symmetry as \textit{negational symmetry}, which is a fundamental property of QNNs because of quantum entanglement.

The main contribution of this work is to introduce and analyze the negational symmetry. We mathematically show this symmetry when the input are binary signals, thus the theoretical results are independent of a particular dataset or model (i.e.~the negational symmetry is independent of the parameters of QNNs), and empirically validate it with a simulated binary pattern classification task. Our experimental results validate that, in contrast to classical NNs, negational symmetry occurs in QNNs. While our study suggests that QNNs could be a new research direction in binary signal processing, our empirical results also show that the negational symmetry is a double-edged sword in practical quantum applications. More specifically, if a binary pattern and its negational counterpart encode different information, a QNN with quantum circuits entangled is not able to separate them. 

Our contributions are fourfold: (1) we formalize, prove, and analyze the negational symmetry of QNNs in quantum binary classification, (2) we propose a representation learning framework for QNNs and generalize the negational symmetry to it; (3) we evaluate the negational symmetry in binary pattern classification on a quantum simulator; (4) we discuss the advantages and disadvantages of negational symmetry in potential applications.

\section{Quantum Binary Classification}
\label{sec:method}
\subsection{Preliminaries}
\label{sec:related}
A variational quantum circuit (VQC) is a quantum circuit model that consists of a set of parametric quantum gates \cite{romero2019variational}. In the near term, a VQC is implemented through the hybrid quantum-classical (HQC) framework. In the HQC framework, a QML task is divided into two subtasks. The first subtask is to apply quantum gates to manipulate qubits in a quantum computer. This quantum process is analogous to the forward pass in a DNN. The second subtask is to optimize the parameters of quantum gates in a classical computer. This classical process is analogous to the backpropagation in a DNN. It has been shown that nonlinear functions can be approximated by VQCs \cite{cao2017quantum,mitarai2018quantum}, which demonstrates the potential values of VQCs in solving practical problems.

\subsection{Problem Formulation}
\label{sec:prob}
\begin{figure}[t]
    \centering
    \includegraphics[width=0.8\linewidth]{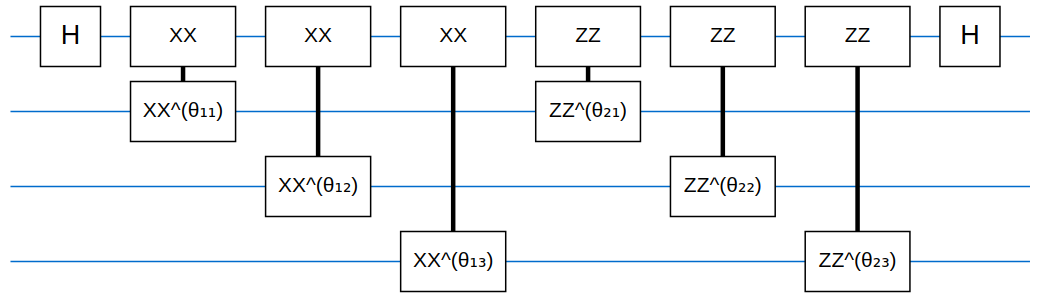}
\caption{The architecture of a 2-layer QNN with 3 + 1 qubits. The input register has 3 data qubits, represented by the second, third and fourth lines from the top. The output register has 1 readout qubit, represented by the top line. The first layer is formed by $X$-parity gates and the second layer is formed by $Z$-parity gates. $\theta_{jk}$ stands for the parameter of the quantum gate operated between the readout qubit and the $k$th data qubit at the $j$th layer. $H$ is the Hadamard gate where $H = \frac{1}{\sqrt{2}} \left[\begin{matrix}
   1 & 1 \\
   1 & -1 
\end{matrix} \right]$. The two $H$ gates denote the pre-processing and post-processing operations on the readout qubit.}
\label{fig:arch}
\end{figure}

In this work, a quantum system is a composite of two systems, namely the \textit{input register} and the \textit{output register}. For an intuitive interpretation, the input register and the output register can be linked with the input data and output data in a ML system, respectively. We use \textit{Dirac} notation~\cite{nielsen2002quantum} to represent quantum data, e.g.~mathematically, $\bra{a}U\ket{b}$ denotes the inner (scalar) product of a row vector $\bra{a}$ and a column vector $\ket{b}$, where $U$ is a linear map.
Given a training dataset $\mathcal{D}_{S} = \{(\ket{\bm{x}_j}, y_j)\}_{j=1}^n$, $\ket{\bm{x}} = \ket{x_1} \otimes \ket{x_2} \cdots \otimes \ket{x_N}$ is a $N$-qubit quantum state for the input register, where $\ket{x} = \alpha \ket{0} + \beta \ket{1}, \alpha, \beta \in \mathbb{C}, |\alpha|^2 + |\beta|^2 = 1$. $y \in \{-1, 1\}$ is the binary label.\footnote{We define $y$ as an integer in a hybrid quantum-classical (HQC) system. In a pure quantum system, $y$ can also be defined as $\ket{y} \in \{\ket{0}, \ket{1}\}$.} The output register is just a \textit{readout} qubit. We prepare the readout qubit as $\ket{1}$. So the input state of the quantum system is $\ket{1, \bm{x}} = \ket{1} \otimes \ket{\bm{x}}$. The readout qubit is pre-processed and post-processed by a Hadamard gate\footnote{The mathematical definition of Hadamard gate is given in the caption of Fig.~\ref{fig:arch}.}, respectively.

Following previous studies \cite{mcclean2018barren,liu2018differentiable,zoufal2019quantum}, we analyze a family of QNNs where the entanglement exists between the readout qubit and each of the data qubits in this work. We define this status as \textit{full entanglement} for simplicity.
Let the unitary $U_{\bm{\theta}}$ be a QNN with parameters $\bm{\theta}$. As the Hadamard gate is non-parametric, we use $U_{\bm{\theta}}$ to denote the integration of the QNN of interest, and the pre- and post-processing gates on the readout qubit for convenience. A forward pass of the input state $\ket{1, \bm{x}}$ through $U_{\bm{\theta}}$ produces the output state $U_{\bm{\theta}} \ket{1, \bm{x}}$. In the traditional quantum circuit models, only the readout qubit is measured by a Hermitian operator $\mathcal{M}$, which is a quantum observable. We limit our choice of $\mathcal{M}$ within Pauli operators, specifically $\mathcal{M} \in \{X, Y, Z\}$.\footnote{See \ref{sec:app_pauli} for details.} As a standard practice in quantum computing, we use $Z$ measurement as the default measurement in this study. The measurement outcome will be either $-1$ or $1$ with uncertainty. When the output state $U_{\bm{\theta}} \ket{\bm{x}, 1}$ is prepared for multiple times, the prediction is defined as the expectation of the observed measurement outcomes. Formally, we have 
\begin{align}
    f_{\bm{\theta}} (\bm{x}) &= \bra{1, \bm{x}} U^{\dagger}_{\bm{\theta}} | \mathcal{M}_{0} |U_{\bm{\theta}} \ket{1, \bm{x}} 
    \label{eq:pred}\\
    &= \mathrm{tr}(| U_{\bm{\theta}}  \ket{1, \bm{x}} \bra{1, \bm{x}} U^{\dagger}_{\bm{\theta}} | \mathcal{M}_{0})
\end{align}
where $\mathcal{M}_{0}$ denotes the measurement on the readout qubit instead of the whole system,\footnote{Mathematically, the measurement on the whole system should be the tensor product of $N+1$ Pauli operators. A simple example could be $\mathcal{M} \otimes  \prod_{\otimes}^{N} I$, where $\prod_{\otimes}^{N} I = \underbrace{I \otimes I \otimes \cdots \otimes I}_N $.} $-1 \leq f_{\bm{\theta}} (\bm{x}) \leq 1$ is a real number and $\mathrm{tr}(\cdot)$ is the trace. The decision boundary on the space of density matrices is the hyperplane $\mathrm{tr}(| U_{\bm{\theta}} \ket{1, \bm{x}} \bra{1, \bm{x}} U^{\dagger}_{\bm{\theta}} | \mathcal{M}_{0} = \tau$), where we set $\tau = 0$.

If we take $f_{\bm{\theta}} (\bm{x})$ as the logit for $\bm{x}$, together with the label $y$, we can define the loss $\mathcal{L}$
\begin{equation}
    \mathcal{L}(f_{\bm{\theta}}(\bm{x}), y) = \max (0, 1 - y \cdot f_{\bm{\theta}}(\bm{x})).
    \label{eq:loss}
\end{equation}
Considering $-1 \leq f_{\bm{\theta}} (\bm{x}) \leq 1$, we choose the hinge loss over the binary cross-entropy (BCE) loss for convenience and robustness~\cite{tzelepi2020improving}. Indeed, the choice of loss function does not influence the conclusion of this study.\footnote{Another common choice of loss function is the fidelity loss $\mathcal{L}_{fidelity} = 1 - fidelity $. The fidelity is defined as $F(\rho_1, \rho_2) = \mathrm{tr}(\sqrt{\sqrt{\rho_1}\rho_2\sqrt{\rho_1}})^2$, where $\rho_1$ and $\rho_2$ are two density matrices. The fidelity loss gives similar results as the hinge loss and BCE but the state preparation and the backpropagation require extra caution in implementation.}

\subsection{Architecture}
\label{sec:arch}
From the perspective of quantum computing, QNNs are variational quantum circuits (VQCs) constructed by different sets of single-qubit quantum gates along with two-qubit entanglement gates. Based on \textit{ZX-calculus}~\cite{coecke2017picturing}, we can prove that any nonlinear function can be $\epsilon$-approximated\footnote{Given a function $f$ and an approximation function $f^{*}$, we have $|f^{*}(x) - f(x)| < \epsilon$ where $\epsilon > 0$.} with single-qubit parametric $Z$-gates ($R_z(\theta)$) and $X$-gates ($R_x(\theta)$), and two-qubit non-parametric CNOT gates.\footnote{See \ref{sec:app_zx} for a sketch of proof.} In this study, to illustrate the impact of entanglement, we use $Z$-parity gates ($ZZ^{\theta} = e^{-i \theta \sigma_z \otimes \sigma_z}$) and $X$-parity gates ($XX^{\theta} = e^{-i \theta \sigma_x \otimes \sigma_x}$) alternatively to build full entanglement \cite{farhi2020classification}, where $\theta$ is the parameter that we want to optimize. Two-qubit $ZZ$ or $XX$ interactions are know as \textit{Ising interactions} in statistical mechanics and each block of parity gates can be viewed as a layer in classical NNs. See Figure~\ref{fig:arch} for the illustration of the architecture. Note, any QNN with full entanglement can always be simplified to this type of architecture according to \textit{ZX-calculus}.

\subsection{Optimization}
\label{sec:opt}
In the near term, the number of parameters is limited by the number of qubits, which is the main challenge for most quantum applications. We choose a gradient-based optimization method because gradient-free algorithms cannot scale up to a larger number of parameters in the long term. For a VQC, the mini-batch gradient has an analytic derivation (based on the chain rule) and a numerical implementation (considering the stochasticity of quantum mechanics). This is the most characteristic difference between the optimization of a QNN and a classical NN~\cite{schuld2019eval}. Although the analytic gradient is fast to compute in a classical environment, the numerical gradient is more robust in a noisy real-world quantum computer. In real quantum applications, the gradient can be approximated by using the \textit{parameter shift rule} \cite{crooks2019gradients}. Given an example pair $(\ket{\bm{x}}, y)$, we define the numerical gradient for a scalar parameter $\theta$ as
\begin{equation}
    \nabla_{\theta} \mathcal{L}(f_{\theta}(\bm{x}), y) = \frac{\mathcal{L}(f_{\theta + \frac{\pi}{2}}(\bm{x}), y) - \mathcal{L}(f_{\theta - \frac{\pi}{2}}(\bm{x}), y)}{2}.
    \label{eq:grad}
\end{equation}

\subsection{Measurement On Data Qubits}
\label{sec:measure}
In the context of deep learning, {\em representation learning}, also known as {\em feature learning}, is a rapidly developing area, \textit{with the goal of yielding more abstract and ultimately more useful representations} of the data, as described by \cite{bengio2013representation}. The composition of multiple non-linear transformations of the data has been used to quantitatively and qualitatively understand the black-box of NNs. For example, in CNNs, the feature maps of lower layers tend to catch the similar basic patterns and the feature maps of higher layers are able to extract the semantic information. However, limited by the physical implementation, it is impractical to extract features at any hidden layers of QNNs. Besides, there is a structural difference between QNNs and classical NNs. In classical NNs, the data is fed into the input layer followed by the hidden layers and the output layer sequentially, while the readout qubit is in parallel with the data qubits in QNNs. 

In this work, we propose to use the measurement on the data qubits as the learned representations of the data. We define the learned feature vector of $\ket{\bm{x}}$ as $g_{\bm{\theta}} (\bm{x})$. Similar to Eq.~\ref{eq:pred}, we have
\begin{equation}
    g_{\bm{\theta}} (\bm{x}) = \bra{1, \bm{x}} U^{\dagger}_{\bm{\theta}} | \mathcal{M}_{1, \cdots, N} |U_{\bm{\theta}} \ket{1, \bm{x}}
    \label{eq:feat}
\end{equation}
where $\mathcal{M}_{1, \cdots, N}$ denotes the measurement on the data qubits instead of the whole system for simplicity. The Hilbert space of the input data $\bm{x}$ is $\mathbb{C}^{2^N}$. So we learn a mapping function $g_{\bm{\theta}}: \mathbb{C}^{2^N} \mapsto [-1, 1]^N$, which projects a quantum state to a real feature vector through transformation in a complex Hilbert space and quantum measurement. It is hard to study the Hilbert space directly. Given two different quantum states $\ket{\bm{x}_j}$ and $\ket{\bm{x}_k}$, we can define Euclidean distance between two feature vectors using the Frobenius norm $||g_{\bm{\theta}} (\bm{x}_j) - g_{\bm{\theta}} (\bm{x}_k)||$ and analyze the representations of quantum output in a classical fashion.

\section{Negational Symmetry of Quantum Neural Networks}
\label{sec:theory}

\subsection{Negational Symmetry for Binary Classification}
\label{sec:theory_neg}
Let us first examine the quantum binary classification with an arbitrary example $\ket{\bm{x}}$. Let $\ket{\bm{x}} = \ket{x_1} \otimes \ket{x_2} \cdots \otimes \ket{x_N}$ be the data qubits of the binary pattern, where $x_i \in \{0, 1\}$ for $i \in \{1, 2, \cdots, N\}$. Then, the inverted binary pattern or the negational counterpart $\ket{\tilde{\bm{x}}} = \ket{\tilde{x}_1} \otimes \ket{\tilde{x}_2} \cdots \otimes \ket{\tilde{x}_N}$, where $\ket{\tilde{x}_i} = X \ket{x_i}$ for $i \in \{1, 2, \cdots, N\}$. Here, quantum gate $X$ is equivalent to bitwise \texttt{NOT} in classical computing. Let us denote $ \bm{X} = \prod_{\otimes}^{N} X$ for simplicity, then we have $\ket{\tilde{\bm{x}}} = \bm{X} \ket{\bm{x}}$. Formally, we introduce the following theorem.

\begin{theorem}
Given a QNN with fixed parameters $\bm{\theta}$ and $Z$-measurement on the readout qubit, $f_{\bm{\theta}} (\bm{x}) = f_{\bm{\theta}} (\tilde{\bm{x}})$.
\label{ob:1}
\end{theorem}

Here, the QNN should have full entanglement, as discussed in Section~\ref{sec:arch}. Since all quantum gates in Eq.~\ref{eq:pred} are 2D matrices, i.e.~linear transformations, the mathematical proof is straightforward and can be found in \ref{sec:app_thm1}. Note, Theorem~\ref{ob:1} describes the situation in expectation due to the nature of measurement operation. Although Eq.~\ref{eq:pred} is a closed-form mathematical expression, in a real quantum device, the empirical observation of $f_{\bm{\theta}} (\bm{x})$ for a single example is dependent on the average of the observed outcomes in repeated measurements, i.e.~$\mathcal{M}_{0}$ is measured in multiple copies. That is to say, Theorem~\ref{ob:1} may not be observed based on a single observation due to the stochasticity of quantum computing. This differs from most classical ML models which have a deterministic output in the inference phase and increases the computational cost in contrast to classical systems.

It is worth mentioning that Theorem~\ref{ob:1} holds not only when a QNN is trained to convergence, but also for a QNN with randomly initiated $\bm{\theta}$. In a 2D Cartesian system, given a function $f$ and a variable $x$ , $f$ is (reflectionally) symmetric if $f(x) = f(-x)$. Similarly, we define Theorem~\ref{ob:1} as the \textit{negational symmetry} for quantum binary classification as there is a symmetry in the measurement on the readout qubit.

To better understand the negational symmetry of QNNs, we decompose a QNN into blocks, as defined in Section~\ref{sec:arch}. We choose the block as the basic unit because each data qubit is entangled with the readout qubit in a block. We study the relationship between the blocks ($ZZ$ block or $XX$ block) and the Pauli measurement ($\{X, Y, Z\}$). The results are summarized in Table~\ref{tab:sym}. Note, as defined in Section~\ref{sec:prob}, we have $-1 \leq f_{\bm{\theta}} (\bm{x}) \leq 1$. The negational symmetry is a built-in property of QNNs when there is at least one $ZZ$ block in a QNN for binary pattern classification.

\begin{table}[t]
    \centering{
    \setlength{\tabcolsep}{0.5em}
    \scriptsize{
    \begin{tabular}{c|c|l}
    \hline
    Architecture & $\mathcal{M}$ & ~~~~~~~~~~~~Symmetry \\ \Xhline{4\arrayrulewidth}
    ($XX$) & Z & $f_{\bm{\theta}} (\bm{x}) = \phantom{-}f_{\bm{\theta}} (\tilde{\bm{x}}) = -\bm{1}$ \\ \hline
    ($XX$) & X & $f_{\bm{\theta}} (\bm{x}) = \phantom{-}f_{\bm{\theta}} (\tilde{\bm{x}}) = \phantom{-}\bm{0}$ \\ \hline
    ($XX$) & Y & $f_{\bm{\theta}} (\bm{x}) = - f_{\bm{\theta}} (\tilde{\bm{x}}) = \phantom{-}\bm{0}$ \\ \hline
    ($ZZ$) & Z & $f_{\bm{\theta}} (\bm{x}) = \phantom{-} f_{\bm{\theta}} (\tilde{\bm{x}})$ \\ \hline
    ($ZZ$) & X & $f_{\bm{\theta}} (\bm{x}) = \phantom{-} f_{\bm{\theta}} (\tilde{\bm{x}}) = \phantom{-}\bm{0}$ \\ \hline
    ($ZZ$) & Y & $f_{\bm{\theta}} (\bm{x}) = - f_{\bm{\theta}} (\tilde{\bm{x}})$ \\ \hline
    ($XX-ZZ$) & Z & $f_{\bm{\theta}} (\bm{x}) = \phantom{-} f_{\bm{\theta}} (\tilde{\bm{x}})$ \\ \hline
    ($XX-ZZ$) & X & $f_{\bm{\theta}} (\bm{x}) = \phantom{-} f_{\bm{\theta}} (\tilde{\bm{x}}) = \phantom{-}\bm{0}$ \\ \hline
    ($XX-ZZ$) & Y & $f_{\bm{\theta}} (\bm{x}) = - f_{\bm{\theta}} (\tilde{\bm{x}})$ \\ \hline
    ($ZZ-XX$) & Z & $f_{\bm{\theta}} (\bm{x}) = \phantom{-} f_{\bm{\theta}} (\tilde{\bm{x}})$ \\ \hline
    ($ZZ-XX$) & X & $f_{\bm{\theta}} (\bm{x}) = \phantom{-} f_{\bm{\theta}} (\tilde{\bm{x}}) = \phantom{-}\bm{0}$ \\ \hline
    ($ZZ-XX$) & Y & $f_{\bm{\theta}} (\bm{x}) = - f_{\bm{\theta}} (\tilde{\bm{x}})$ \\ \hline
    \end{tabular}
    }
    \caption{A block-wise study of the negational symmetry for binary classification.}
    \label{tab:sym}
    }
\end{table}

\subsection{Negational Symmetry for Representation Learning}
\label{sec:neg_feat}
\begin{table}[t]
  \centering{
    \captionsetup{type=table}
    \setlength{\tabcolsep}{0.5em}
    \scriptsize{
    \begin{tabular}{c|l}
    \hline
    $\mathcal{M}$ & ~~~~~~~~Symmetry \\ \Xhline{4\arrayrulewidth}
    $Z$ & $g_{\bm{\theta}} (\bm{x}) = - g_{\bm{\theta}} (\tilde{\bm{x}})$ \\ \hline
    $X$ & $g_{\bm{\theta}} (\bm{x}) = \phantom{-} g_{\bm{\theta}} (\tilde{\bm{x}}) = \bm{0}$ \\ \hline
    $Y$ & $g_{\bm{\theta}} (\bm{x}) = - g_{\bm{\theta}} (\tilde{\bm{x}})$ \\ \hline
    \end{tabular}
    }
    \caption{Negational symmetry of representations for Pauli measurement.}
    \label{tab:sim}
    }
\end{table}
Analogous to classical representation learning, we need a tool to analyze the learned representations of QNNs. Here, we propose to use the expectation of the observed measurement outcomes as the learned representations and perform the same analysis with the mathematical tools that we use for classical NNs. As described in Section~\ref{sec:measure}, we measure $N$ data qubits with Pauli measurement ($\{X, Y, Z\}$). The output $g_{\bm{\theta}} (\bm{x})$ is a $N$-element feature vector in the real domain. Given the real feature vectors, we can use statistical tools to study the relationship between feature vectors, e.g.~Pearson's correlation coefficient and cosine similarity. Besides, we can visualize the features for qualitative comparison. For example, we can visualize the learned representations with t-Distributed Stochastic Neighbor Embedding (t-SNE) \cite{maaten2008visualizing}. Note, although we can visualize the final representations of a QNN, we cannot access the intermediate representations in the tensor product Hilbert space directly.

In addition to the symmetry in the measurement on the readout qubit, there is also a symmetry in the learned representations. Given the same assumptions in Theorem~\ref{ob:1}, we have:
\begin{theorem}
Given a QNN with fixed parameters $\bm{\theta}$ and $Z$-measurement on the data qubits, $g_{\bm{\theta}} (\bm{x}) = -g_{\bm{\theta}} (\tilde{\bm{x}})$.
\label{ob:2}
\end{theorem}
Compared with Theorem~\ref{ob:1}, Theorem~\ref{ob:2} is more intuitive, where the feature vectors of the binary pattern and its negational counterpart have opposite directions in the feature space. The mathematical relationship between $g_{\bm{\theta}} (\bm{x})$ and $g_{\bm{\theta}} (\tilde{\bm{x}})$ is analogous to rotational symmetry (symmetric to the origin) in a Cartesian coordinate system. We define Theorem~\ref{ob:2} as the \textit{negational symmetry} of quantum representation learning. Again, we summarize the symmetry for all Pauli measurement in Table~\ref{tab:sim}. 

\section{Experiments}
The purpose of the simulated experiments in this section are twofold. First, we want to validate our theoretical findings in Sec.~\ref{sec:theory}. Second, we want to demonstrate the downside of negational symmetry, where will show a practical ML problem that QNNs cannot solve.

\label{sec:exp}
\subsection{Experimental Setting}
\paragraph{Environment}
All experiments were run on a classical computer with Ubuntu 18.04 LTS.\footnote{The code and pre-trained weights are publicly available at \url{https://github.com/eveningdong/negational_symmetry}.} We simulate the NISQ circuits with \texttt{Cirq}.\footnote{Cirq is the main research tool used by the Google AI Quantum team. \url{https://quantumai.google}} The CPU is an Intel\textsuperscript{\textregistered} Xeon\textsuperscript{\textregistered} Processor E5-2686 v4 $@$ 2.30 GHz with 45 MB cache. The GPU is a NVIDIA\textsuperscript{\textregistered} Tesla\textsuperscript{\textregistered} V100 with 16 GB memory. The RAM is up to 64 GB. Note, while simulators allow us to simulate QNNs on classical computers, this simulation does not scale up to a large number of qubits. As the Hilbert space of the input data $\mathbb{C}^{2^N}$ grows exponentially with the increase of $N$, the classical computers easily reach their memory limit to simulate the quantum process.

\paragraph{Hyperparameters}
We use the same set of hyperparameters for the training of classical NNs. We use Adam \cite{kingma2015adam}, a gradient-based stochastic optimization method , with $\beta_1 =0.9$, $\beta_2=0.999$, and $\epsilon=10^{-7}$. The batch size is 32. The constant learning rate is $10^{-4}$. We train all models to converge.

\paragraph{Dataset}
Following \cite{bausch2020recurrent,farhi2020classification}, we use the binarized MNIST dataset for binary classification tasks and we generate binary patterns with digits 3 and 6. At the beginning, the training set contains 6131 images labeled as 3 and 5918 images labeled as 6. Limited by the hardware and following  \cite{farhi2020classification}, the images are downsampled from $28 \times 28$ to $4 \times 4$ to fit the simulator, but the results generalize to any size of qubits as the result of Theorem~\ref{ob:1}. We then map each grayscale pixel value to $\{0, 1\}$ with 128 as the threshold and remove the contradictory examples (the images labeled as both 3 and 6 simultaneously). Here, $\{0, 1\}$ is equivalent to black-and-white image classification in computer vision. Then, in the quantum state preparation step, $\{0, 1\}$ is mapped to $\{\ket{0}, \ket{1}\}$ for each qubit. After the preprocessing, the training set consists of 3649 images while there are 2074 images labeled as 3 and 1575 images labeled as 6. With the same procedure, the final test set consists of 890 images while there are 332 images labeled as 3 and 558 images labeled as 6. The images are flatten into vectors.

\subsection{Negational Symmetry in Binary Pattern Classification}
\label{sec:exp_mnist}
\begin{figure}[t]
    \centering
    \begin{subfigure}[t]{0.32\columnwidth}
        \centering
        \includegraphics[width=\linewidth]{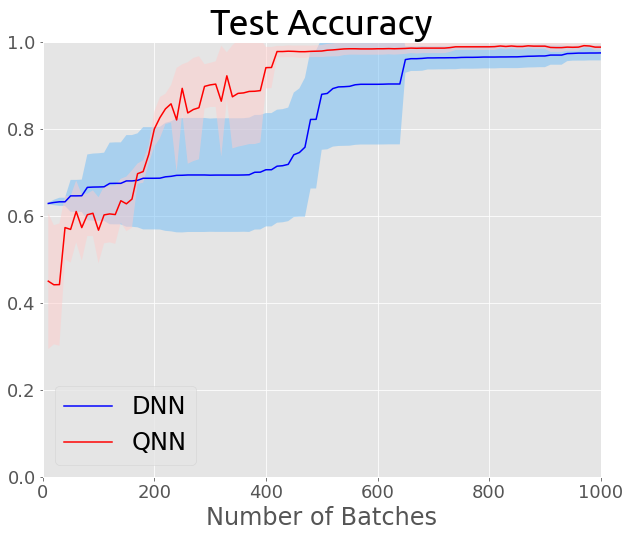}
        \caption{Test accuracy on the test set.}
    \end{subfigure}
    \begin{subfigure}[t]{0.32\columnwidth}
        \centering
        \includegraphics[width=\linewidth]{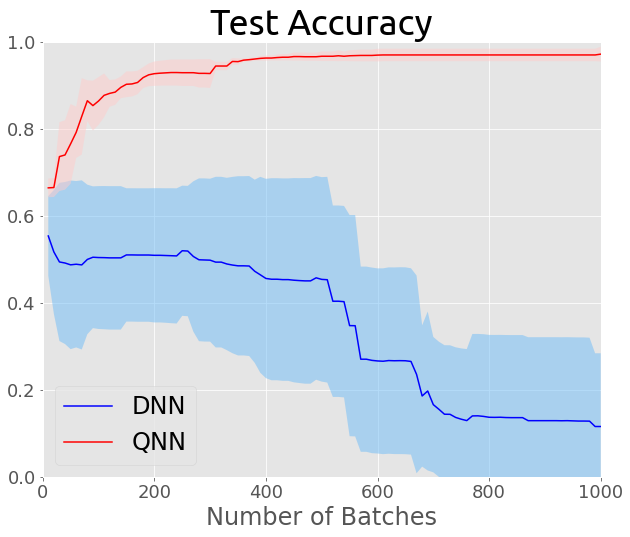}
        \caption{Test accuracy on the negational test set.}
    \end{subfigure}
\caption{Comparison between the DNN and the QNN on MNIST. The shaded region is 1 standard deviation over 5 runs with different random seeds.}
\label{fig:mnist}
\end{figure}

\begin{figure}[t]
    \centering
    \begin{subfigure}[t]{0.16\columnwidth}
        \centering
        \includegraphics[width=\linewidth]{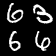}
        \caption{}
    \end{subfigure}
    \begin{subfigure}[t]{0.16\columnwidth}
        \centering
        \includegraphics[width=\linewidth]{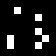}
        \caption{}
    \end{subfigure}
    \begin{subfigure}[t]{0.16\columnwidth}
        \centering
        \includegraphics[width=\linewidth]{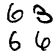}
        \caption{}
    \end{subfigure}
    \begin{subfigure}[t]{0.16\columnwidth}
        \centering
        \includegraphics[width=\linewidth]{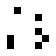}
        \caption{}
    \end{subfigure}
    \caption{Visualization: (a) the original test images; (b) the binarized test images; (c) the inverted test images; (d) the binarized inverted test images.}
    \label{fig:mnist_vis}
\end{figure}

We numerically validate Theorem~\ref{ob:1} by evaluating the negational symmetry of QNNs in simulated binary pattern classification tasks. We use the 2-layer QNN described in Section~\ref{sec:arch}. The readout qubit is measured by a Pauli $Z$ operator. The QNN has a 16-qubit input register and 32 parameters in total. The test results are presented in Figure~\ref{fig:mnist}(a). We also repeat the above experiment in the negational setting. This time, we invert the grayscale MNIST images of the test set before the pre-processing step. From a classical view, we exchange the colors of the pixels of the digit (white to black) and the pixels of the background (black to white). See Figure~\ref{fig:mnist_vis} for the intuition. Simply, the data qubits that were in the state $\ket{0}$ are now in the state $\ket{1}$, and vice versa. This bit-flipping operation is achieved via an $X$ gate. In fact, this negational (or bit-flipping) operation creates a \textit{domain shift} \cite{shimodaira2000improving} if we consider the training set as the source domain and the test set as the target domain from the perspective of classical ML. That is to say, we have a transfer learning problem as we want to extract knowledge learned from the training set but apply it to the negational test set. Given Theorem~\ref{ob:1}, it is not surprising that QNNs should continue to maintain a high performance on the test set with negational operation. The test results are presented in Figure~\ref{fig:mnist}(b).

To validate the universality of negational symmetry for quantum binary classification, we evaluate QNNs with different architectures under the same experimental setting. We first extend the 2-layer QNN (\textit{XX-ZZ}, denoted as $Q_1$) to deeper QNNs, namely a 3-layer QNN (\textit{XX-ZZ-XX}, denoted as $Q_2$) and a 4-layer QNN (\textit{XX-ZZ-XX-ZZ}, denoted as $Q_3$), to study the effect of the depth on model performance. We then study the order of the blocks of $X$-parity gates and $Z$-parity gates with a 2-layer QNN (\textit{XX-ZZ}, denoted as $Q_4$) and 3-layer QNN (\textit{XX-ZZ-XX}, denoted as $Q_5$). For these larger models, we use early stopping to report best test accuracy. We train the models on the training set with the \textbf{same} random seeds, freeze the weights and evaluate them on the test set without negational operation and the test set with negational operation. The results are present in Table~\ref{tab:qnn}. As expected, we observe for each of the QNNs the same performance on the test sets with and without negation. The same experiments were repeated for different two-digit combinations and the same phenomena above were observed. Note, as the negational symmetry is an inherent property of the model, it is independent of the learned parameters and also occurs in a randomly initialized model.

\begin{table}[t]
    \centering{
    \scriptsize{
    \begin{tabular}{lrrr}
    \toprule
    Model & w/o Negation & w/ Negation & \# Params \\
    \midrule
    $Q_1$ & 0.9783 & 0.9783 & 32 \\ 
    $Q_2$ & 0.9674 & 0.9674 & 48 \\ 
    $Q_3$ & 0.9922 & 0.9922 & 64 \\ 
    $Q_4$ & 0.9707 & 0.9707 & 32 \\
    $Q_5$ & 0.9967 & 0.9967 & 48 \\
    \bottomrule
    \end{tabular}
    }
    \caption{Evaluation of QNNs with different architectures on negational symmetry.}
    \label{tab:qnn}
    }
\end{table}

We further compute the difference between the logits of the QNNs, i.e.~Eq.~\ref{eq:pred}, for all $890$ pairs in the two test sets. We find that the numerical difference is negligible. For example, the mean of the differences is $-4.4240963 \times 10^{-8}$ and the standard deviation is $1.05649356 \times 10^{-7}$ for $Q_1$. If we take the noisy environment of the NISQ device and the number of significant figures into account, we can say that QNNs make the same prediction on a binary pattern and its negational counterpart.

To contrast this with classical NNs, we repeat the experiment with a 2-layer DNN whose number of parameters is close to the QNN's number of parameters. The first layer of the DNN is a fully-connected layer with 16 input nodes and 2 output nodes, followed by a ReLU activation function \cite{nair2010relu}. The second layer is a fully-connected (FC) layer with 2 input nodes and 1 output node. The total number of parameters for the DNN is 37. We use BCE as the loss function for DNNs. The results are presented in Figure~\ref{fig:mnist}. Both the DNN and the QNN achieve promising results in the first scenario on the test set without negation. The QNN seems to converge faster and more stable than the DNN, when the training does not suffer from \textit{barren plateaux} \cite{mcclean2018barren}. However, the DNN fails in the negational setting. Additionally, we provide the results of 4 DNNs with different number of hidden nodes and different number of hidden layers (denoted as $D_{i=\{1,2,3,4\}}$) and CNNs including a 2-layer CNN $C_1$ and a 3-layer CNN $C_2$ in Table~\ref{tab:nn}.\footnote{See \ref{sec:app_exp} for the details of the model architecture.} For classical NNs, more parameters and advanced architectures improve the performance on the original test set, but do not help on the negational one.

\begin{table}[ht]
    \centering{
    \scriptsize{
    \begin{tabular}{lrrr}
    \toprule
    Model & w/o Negation & w/ Negation & \# Params\\
    \midrule
    $Q_1$ & 0.9783 & 0.9783 & 32 \\ 
    $D_1$ & 0.9775 & 0.1146 & 37 \\
    $D_2$ & 0.9944 & 0.1528 & 289 \\
    $D_3$ & 0.9978 & 0.0067 & 309 \\ 
    $D_4$ & 0.9933 & 0.0090 & 561 \\
    $C_1$ & 0.9978 & 0.3079 & 701 \\
    $C_2$ & 0.9978 & 0.0247 & 1585 \\
    \bottomrule
    \end{tabular}
    }
    \caption{Comparison of QNN and classical NNs on negational symmetry. DNNs and CNNs do not have negational symmetry.}
    \label{tab:nn}
    }
\end{table}

\subsection{Negational Symmetry in Quantum Representation Learning}
\begin{figure}[t]
    \centering
    \begin{subfigure}[t]{0.4\columnwidth}
        \centering
        \includegraphics[width=\linewidth]{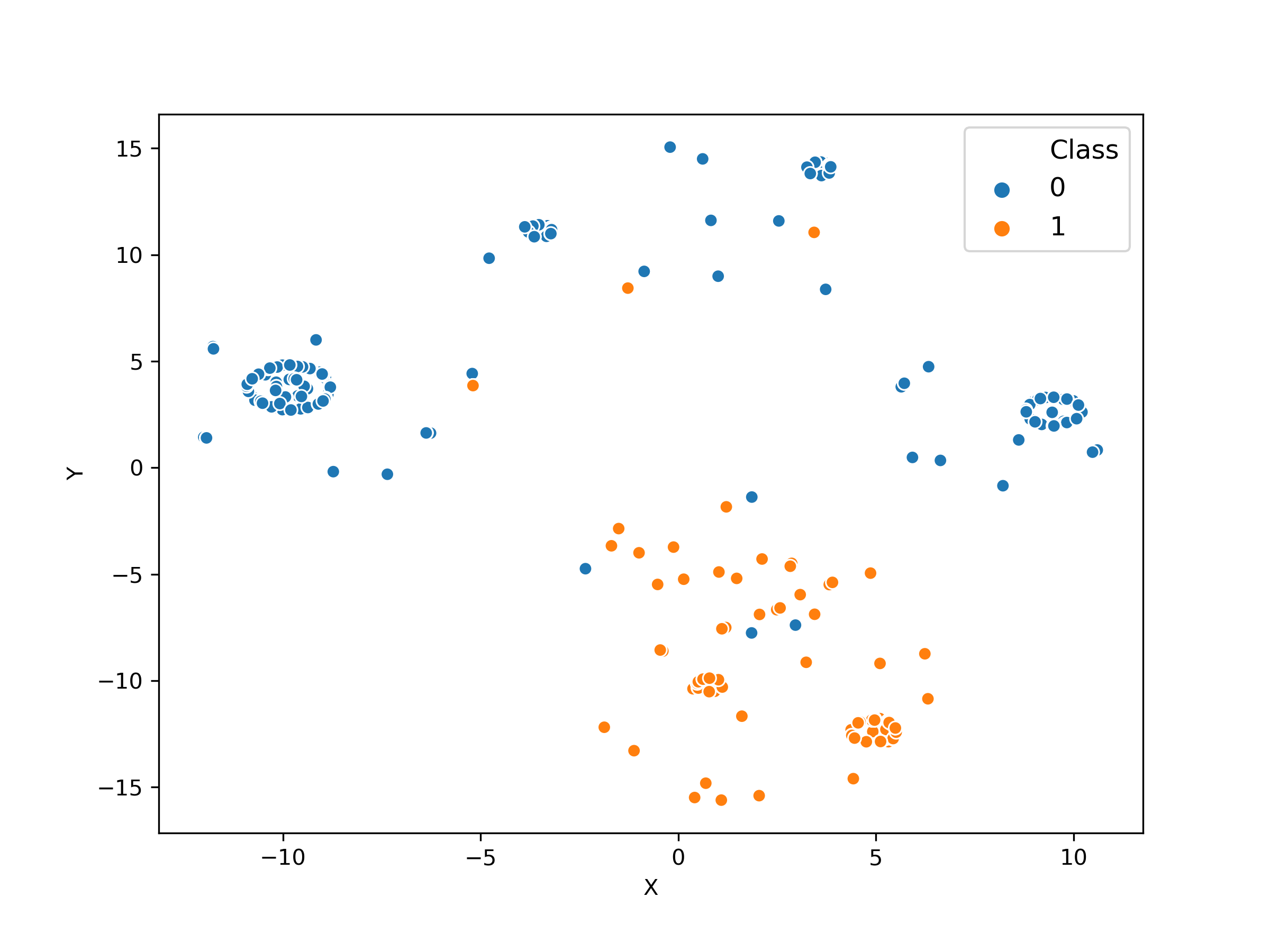}
        \caption{$Z$ measurement on the test set w/o negational operation.}
    \end{subfigure}
    \begin{subfigure}[t]{0.4\columnwidth}
        \centering
        \includegraphics[width=\linewidth]{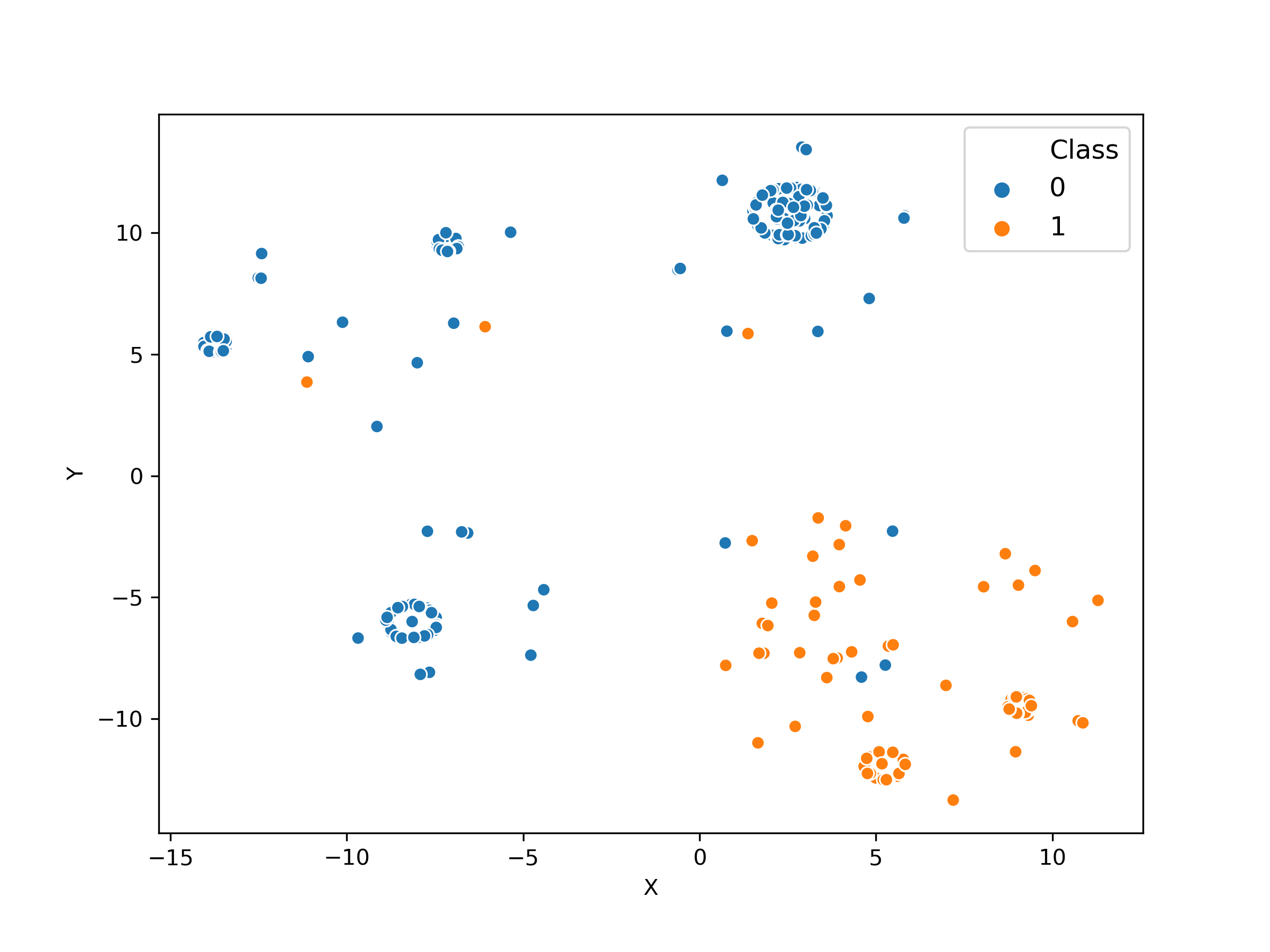}
        \caption{$Z$ measurement on the test set w/ negational operation.}
    \end{subfigure}
\caption{t-SNE visualization of learned representations.}
\label{fig:tsne}
\end{figure}

For Theorem~\ref{ob:2}, we first calculate the norm of the sum of two pairwise feature vectors $||g_{\bm{\theta}} (\bm{x}) + g_{\bm{\theta}} (\tilde{\bm{x}})||$\footnote{Note, we use $|$ in \emph{Dirac} notations and use $||$ in norm operations.}. The sum of the norms is less than $1 \times 10^{-6}$, which numerically validates the negational symmetry in the auxiliary quantum representation learning. Following the discussion in Section~\ref{sec:neg_feat}, we also check the pairwise statistical similarity for the feature vectors in the original test set and its negational counterpart. For $Z$ measurement, the mean for pairwise Pearson's correlation coefficient is $-0.5$ and the mean for pairwise cosine similarity is $-1$. Additionally, we provide the t-SNE visualizations with the \textbf{same} random seed in Figure~\ref{fig:tsne}. The clusters formed by the learned representations of the two classes can be clearly split, which suggests that the measurement on the data qubits could be a meaningful tool for quantum representation learning.

\subsection{Drawback of Negational Symmetry}
In the above experiments, we validate that QNNs cannot differentiate a binary pattern and its negational counterpart, and highlight its advantage when the two patterns encode the same semantic information (i.e.~the digit). However, if the two patterns actually encode different semantic information, negational symmetry of QNNs could be hazardous to the system. Here, we provide another simple experiment where both QNNs and NNs are expected to learn to differentiate a binary pattern and its negational counterpart. We use the same networks $Q_1$, $D_1$, and $C_1$ as in Section~\ref{sec:exp_mnist}. We use the same training set and test set in the above experiments. However, we include both original patterns and the corresponding negational patterns in both training set and test set this time. Instead of using the original digit labels, the original patterns are assigned the label $1$ (or $\ket{1}$) and the negational patterns are assigned the label $0$ (or $\ket{0}$). Thus, the objective becomes to separate the white digits on the black background from the black digits on the white background. The results are present in Table~\ref{tab:drawback}. As expected, based on Theorem~\ref{ob:1}, QNNs are not able to solve this simple task and achieve an accuracy of around $50\%$ (random guess), while the classical models perfectly solve the task. It is thus crucial to note that negational symmetry in practical settings could be a double-edged sword: when a binary pattern and its negational counterpart encode different information, a system built on QNNs could be vulnerable to malicious attacks.

\begin{table}[ht]
    \centering{
    \scriptsize{
    \begin{tabular}{lcc}
    \toprule
    Model & Accuracy & \# Params\\
    \midrule
    $Q_1$ & 0.5007 $\pm$ 0.0003 & \phantom{0}32 \\ 
    $D_1$ & 1.0000 $\pm$ 0.0000 & \phantom{0}37 \\
    $C_1$ & 1.0000 $\pm$ 0.0000 & 701 \\
    \bottomrule
    \end{tabular}
    }
    \caption{Comparison of QNN and classical NNs on learning to identify the negation operation on the binary patterns. QNNs fail because of negational symmetry.}
    \label{tab:drawback}
    }
\end{table}

\section{Limitations}
\label{sec:limit}
We only study the negational symmetry of QNNs for quantum input in \textit{pure} states (e.g.~$\{\ket{0}, \ket{1}\}$). It will also be interesting to generalize the theoretical analysis to \textit{mixed} states\footnote{$\ket{x} = \alpha \ket{0} + \beta \ket{1}, \alpha, \beta \in \mathbb{C}, |\alpha|^2 + |\beta|^2 = 1$} in the future. The practical application of QNNs may require further discussion. In fact, we can only afford a GPU-based simulated environment with 16 qubits. Limited by the number of qubits, we only have 16-bit binary patterns following \cite{farhi2020classification}, which limits the experimental design. In the long term, we expect more advanced mathematical and experimental tools to emerge from new joint developments in mathematics, physics, and engineering.

It is worth mentioning that this study mainly discusses on the theoretical property of QNNs in binary pattern classification, while the quantum state preparation, as a physical operation, is beyond the scope of this work. In fact, the step of quantum state preparation could easily hide the computational complexity of quantum models~\cite{bang2019optimal}, which also limits the scalability of quantum models to handle complex tasks (e.g.~it is still challenging to deploy large-scale qubits in a quantum computer~\cite{arute2019quantum}). However, we expect that the quantum state preparation will not be a bottleneck for QNNs, as the development of quantum hardware progresses.

\section{Conclusions}
\label{sec:con}
In this work, we present the negational symmetry of QNNs in binary pattern classification, a fundamental property of QNNs that has not been observed previously. We formalize and prove this property and discuss the mechanisms behind. We empirically validate the existence of this new form of symmetry that is inherent in QNNs by simulated experiments and demonstrate that negational symmetry could be a double-edged sword in quantum applications. For pattern recognition research in quantum devices in the long run, we believe that a better theoretical understanding of the properties of QNNs are required and that properties such as the negational symmetry need to be taken into account when designing QNNs to solve practical problems.

\section*{Acknowledgment}
We would like to thank Bob Coecke and Aleks Kissinger from the Department of Computer Science, University of Oxford, Shuxiang Cao from the Department of Physics, University of Oxford, and Edward Grant from the Department of Computer Science, University College London for valuable discussion. We would also like to thank Amazon, Google and Huawei for providing GPU computing service for this study.

\bibliographystyle{elsarticle-num}
\bibliography{refs}

\appendix
\section{Quantum Information Basics}
\label{sec:app_basic}

\subsection{Qubit}
In quantum computing, the basic unit of information is a quantum bit or qubit. A qubit can be realized by different physical systems with two perfectly distinguishable states, e.g. the vertical polarization and horizontal polarization of a single photon. Assume each qubit is in one of two perfectly distinguishable states, we can represent the binary pattern by qubits.

\subsection{Tensor Product Hilbert Space}
For system $A$ with Hilbert space $\mathbb{H}_A = \mathbb{C}^{d_A}$ with dimension $d_A$ and system $B$ with Hilbert space $\mathbb{H}_B = = \mathbb{C}^{d_B}$ with dimension $d_B$, the Hilbert space of the composite system $AB$ is the tensor product of the Hilbert spaces of $A$ and $B$. In formula, $\mathbb{H}_{AB} = \mathbb{H}_A \otimes \mathbb{H}_B = \mathbb{C}^{d_A d_B}$.

\subsection{Pauli Matrices}
\label{sec:app_pauli}
\begin{equation}
   \begin{split}
  \sigma_0 = I\,=\,\left[ \begin{matrix}
   1 & 0  \\
   0 & 1  \\
\end{matrix} \right] ,
  \sigma_x = X = \left[ \begin{matrix}
   0 & 1  \\
   1 & 0  \\
\end{matrix} \right] , \\
  \sigma_y = Y = \left[ \begin{matrix}
   0 & -i  \\
   i & 0  \\
\end{matrix} \right] ,
  \sigma_z = Z = \left[ \begin{matrix}
   1 & 0  \\
   0 & -1  \\
\end{matrix} \right] .
\end{split} 
\end{equation}

\begin{equation}
  \sigma_0^2 = \sigma_x^2 = \sigma_y^2 = \sigma_z^2 = I
\end{equation}

\begin{align}
  X\ket{0} = \ket{1}, X\ket{1} = \ket{0}, \nonumber \\
  Y\ket{0} = i\ket{1}, Y\ket{1} = i\ket{0},  \\
  Z\ket{0} = \ket{0}, Z\ket{1} = -\ket{1}  \nonumber 
\end{align}

\subsection{Pauli Rotation Operators}
\label{sec:app_pauli_op}
The rotation operators are generated by exponentiation of the Pauli matrices according to $e^{(i A \theta)} = \cos(\theta) I + i \sin (\theta) \mathcal{M}$, where $\mathcal{M} \in \{X, Y, Z\}$. The rotation gate $R_a(\theta)$ is a single-qubit rotation through angle $\theta$ (radians) around the corresponding axis $a \in \{x, y, z\}$.

\begin{equation}
    \begin{split}
    \label{R_x}
       {{R}_{x}}\left( \theta  \right) &={{e}^{-i\frac{\theta X}{2}}}=\cos \left( \frac{\theta }{2} \right)I-i\sin \left( \frac{\theta }{2} \right)X =\left[ \begin{matrix}
   \cos \left( \frac{\theta }{2} \right) & -i\sin \left( \frac{\theta }{2} \right)  \\
   -i\sin \left( \frac{\theta }{2} \right) & \cos \left( \frac{\theta }{2} \right)  \\
   \end{matrix} \right]\, 
   \end{split}
\end{equation}

\begin{equation}
\begin{split}
 \label{R_y}
    {{R}_{y}}\left( \theta  \right)& ={{e}^{-i\frac{\theta \text{Y}}{2}}}=\cos \left( \frac{\theta }{2} \right)I-i\sin \left( \frac{\theta }{2} \right)Y =\left[ \begin{matrix}
    \cos \left( \frac{\theta }{2} \right) & -\sin \left( \frac{\theta }{2} \right)  \\
    \sin \left( \frac{\theta }{2} \right) & \cos \left( \frac{\theta }{2} \right)  \\
\end{matrix} \right] 
\end{split}
\end{equation}
  
\begin{equation}
    \begin{split}
       \label{R_z}
    {{R}_{z}}\left( \theta  \right) &={{e}^{-i\frac{\theta Z}{2}}}=\cos \left( \frac{\theta }{2} \right)I-i\sin \left( \frac{\theta }{2} \right)Z =\left[ \begin{matrix}
   {{e}^{-i\frac{\theta }{2}}} & 0  \\
   0 & {{e}^{i\frac{\theta }{2}}}  \\
\end{matrix} \right] 
    \end{split}
\end{equation}

\subsection{Universality of ZX-calculus}
\label{sec:app_zx}
According to Eq.~(\ref{R_x}), Eq.~(\ref{R_y}), and Eq.~(\ref{R_z}), we notice that $R_y(\theta) = Z^{\frac{1}{2}} R_x(\theta) Z^{\frac{1}{2} \dagger}$, which we can use $R_x$ and $R_z$ to represent $R_y$ with arbitrary angles. Formally, we have the following theorem.

\begin{theorem}[\cite{coecke2017picturing}]
\label{thm:euler}
For any unitary $U$ on a single qubit there exist phases $\alpha$, $\beta$, and $\gamma$ such that $U$ can be written as: $U = R_x(\gamma) R_z(\beta) R_x(\alpha)$. This is called the Euler decomposition of $U$ and the phases $\alpha$, $\beta$, and $\gamma$ are called the Euler angles.
\end{theorem}

\begin{theorem}[\cite{coecke2017picturing}]
\label{thm:nqubit}
Any n-qubit unitary can be constructed out of the CNOT gate and phase gates.
\end{theorem}

\begin{theorem}
\label{thm:zx}
For any nonlinear function, there exists at least one $\mathrm{VQC}$ $U_{\bm{\theta}}$ with following properties: (1) it can be constructed out of the gate set $\mathbb{U} = \{R_x, R_z, \mathrm{CNOT}\}$ with parameters $\bm{\theta}$; (2) it can $\epsilon$-approximate the nonlinear function for $\epsilon > 0$.
\proof{
$R_x$ and $R_z$ are also called \textit{phase gates} in ZX-calculus. Here, we use the fact that single-qubit elementary gates and two-qubit gates such as $XX, ZZ, CZ$ are special cases of the gates in $\mathbb{U}$ or can be constructed out of $\mathbb{U}$. Theorem~\ref{thm:zx} is a direct result of the Universality Theorem of neural networks and the Universality Theorem of ZX-calculus.
}
\end{theorem}

\subsection{Quantum Correlation}
\label{sec:app_cor}
Assume there is a composite system $AB$, where $A$ and $B$ are two qubits. We have $\ket{\Phi^{+}} = \mathrm{CNOT} (H \otimes I) (\ket{0} \otimes \ket{0}) = \frac{1}{\sqrt{2}} (\ket{0} \otimes \ket{0} + \ket{1} \otimes \ket{1})$, i.e.~we create a Bell state through an entangling gate. If we measure $A$ and $B$ both in the same basis, we can verify that for the composite system $AB$, the outcomes of $A$ and $B$ are perfectly correlated. After the entangling gate, $A$ and $B$ become perfectly correlated. This phenomenon is called \textit{quantum steering} or quantum correlation.

\section{Theoretical Analysis}
\subsection{Sketch of Proof for Theorem 1}
\label{sec:app_thm1}
The mathematical proof for Theorem~1 is straightforward for QNNs with finite qubits. For simplicity, we assume that there is only one readout qubit $\ket{1}$ and one data qubit $\ket{0}$ (the opposite is then $\ket{1}$. Here, we demonstrate the proof for QNN (\textit{XX-ZZ}) with Z-measurement. The proof for multiple data qubits and QNNs with different architectures follow the same logic. We use the notations in Section~2 and Section~3. 

Following Eq.~1, we have 
\begin{equation}
f_{\bm{\theta}} (\bm{x}) = \bra{1, 0} U^{\dagger}_{\bm{\theta}} | Z \otimes I |U_{\bm{\theta}} \ket{1, 0}, 
\label{eq:app_f}
\end{equation}
where 
\begin{equation}
 U_{\bm{\theta}} =
(H \otimes I) (R_{x}(\theta_1) \otimes R_{x}(\theta_1)) (R_{z}(\theta_2) \otimes R_{z}(\theta_2)) (H \otimes I).  
\label{eq:app_U}
\end{equation}
We have
\begin{equation}
H \otimes I = \frac{1}{\sqrt{2}} 
    \left[\begin{matrix}
   1 & 1 & 0 & 0 \\
   1 & -1 & 0 & 0 \\
   0 & 0 & 1 & 1 \\
   0 & 0 & 1 & -1 \\
\end{matrix} \right], \nonumber
\end{equation}
\begin{align}
R_{x}(\theta_1) \otimes R_{x}(\theta_1) = 
    &\left[\begin{matrix}
    \cos(\frac{\theta_1}{2}) & 0 & 0 & -i \sin(\frac{\theta_1}{2}) \\
    0 & \cos(\frac{\theta_1}{2}) & -i \sin(\frac{\theta_1}{2}) & 0 \\
    0 & -i \sin(\frac{\theta_1}{2}) & \cos(\frac{\theta_1}{2}) & 0 \\
    -i \sin(\frac{\theta_1}{2}) & 0 & 0 & \cos(\frac{\theta_1}{2}) \\
\end{matrix} \right], \nonumber
\end{align}
and 
\begin{equation}
R_{z}(\theta_2) \otimes R_{z}(\theta_2) = 
    \left[\begin{matrix}
    e^{-i \theta_2} & 0 & 0 & 0 \\
    0 & 1 & 0 & 0 \\
    0 & 0 & 1 & 0 \\
    0 & 0 & 0 & e^{i \theta_2} \\
\end{matrix} \right]. \nonumber
\end{equation}
Substitute $H \otimes I$, $R_{x}(\theta_1) \otimes R_{x}(\theta_1)$ and $R_{z}(\theta_2) \otimes R_{z}(\theta_2)$ into Eq.~\ref{eq:app_U}, we have $U_{\bm{\theta}}$ in Eq.~\ref{eq:app_U_theta}.

\begin{equation}
{\scriptsize
U_{\bm{\theta}} = \frac{1}{2}\\
\left[\begin{matrix}
    \cos(\frac{\theta_1}{2})(e^{-i \theta_2} + 1) & \cos(\frac{\theta_1}{2}) (e^{i \theta_2} - 1) & -i \sin(\frac{\theta_1}{2})(e^{-i \theta_2} + 1) & i \sin(\frac{\theta_1}{2})(e^{-i \theta_2} - 1) \\
    \cos(\frac{\theta_1}{2}) (e^{-i \theta_2} - 1) & \cos(\frac{\theta_1}{2}) (e^{-i \theta_2} + 1) & -i \sin(\frac{\theta_1}{2})(e^{-i \theta_2} - 1) & i \sin(\frac{\theta_1}{2})(e^{-i \theta_2} + 1) \\
    -i \sin(\frac{\theta_1}{2})(1 + e^{-i \theta_2}) & i \sin(\frac{\theta_1}{2})(1 - e^{-i \theta_2}) & \cos(\frac{\theta_1}{2})(1 + e^{i \theta_2}) & \cos(\frac{\theta_1}{2})(1 - e^{i \theta_2}) \\
    -i \sin(\frac{\theta_1}{2})(1 - e^{-i \theta_2}) & i \sin(\frac{\theta_1}{2})(1 + e^{-i \theta_2}) & \cos(\frac{\theta_1}{2})(1 - e^{i \theta_2}) & \cos(\frac{\theta_1}{2})(1 + e^{i \theta_2}) \\
\end{matrix} \right]
}
\label{eq:app_U_theta}
\end{equation}

We have 
\begin{equation}
\begin{split}
    U_{\bm{\theta}} \ket{1, 0} = 
    \left[\begin{matrix}
    - i \sin(\frac{\theta_1}{2})(e^{-i \theta_2} + 1) \\
    - i \sin(\frac{\theta_1}{2})(e^{-i \theta_2} - 1) \\
    \cos(\frac{\theta_1}{2})(1 + e^{i \theta_2}) \\
    \cos(\frac{\theta_1}{2})(1 - e^{i \theta_2})
\end{matrix} \right] =
\left[\begin{matrix}
   - \sin(\frac{\theta_1}{2}) \sin(\theta_2) - i \sin(\frac{\theta_1}{2})(\cos(\theta_2) + 1) \\
   - \sin(\frac{\theta_1}{2}) \sin(\theta_2) - i \sin(\frac{\theta_1}{2})(\cos(\theta_2) - 1) \\
   \cos(\frac{\theta_1}{2})(1 + \cos(\theta_2)) + i \cos(\frac{\theta_1}{2}) \sin(\theta_2)  \\
   \cos(\frac{\theta_1}{2})(1 - \cos(\theta_2)) - i \cos(\frac{\theta_1}{2}) \sin(\theta_2)  \\
\end{matrix} \right] 
\end{split}
\end{equation}, where 
\begin{equation}
\ket{1, 0} = 
    \left[\begin{matrix}
   0 & 0 & 1 & 0 \\
\end{matrix} \right]^{T}. \nonumber
\end{equation}

Substitute $Z \otimes I$ and $U_{\bm{\theta}} \ket{1, 0}$ into Eq.~\ref{eq:app_f}, we have
\begin{equation}
f_{\bm{\theta}} (\bm{x}) = \sin^2(\frac{\theta_1}{2}) - \cos^2(\frac{\theta_1}{2}),
\end{equation}
where 
\begin{equation}
Z \otimes I = 
    \left[\begin{matrix}
   1 & 0 & 0 & 0 \\
   0 & 1 & 0 & 0 \\
   0 & 0 & -1 & 0 \\
   0 & 0 & 0 & -1 \\
\end{matrix} \right]. \nonumber
\end{equation}

Similarly, we have 
\begin{equation}
    \begin{split}
    U_{\bm{\theta}} \ket{1, 1} = 
    \left[\begin{matrix}
    i \sin(\frac{\theta_1}{2})(e^{-i \theta_2} - 1) \\
    i \sin(\frac{\theta_1}{2})(e^{-i \theta_2} + 1) \\
    \cos(\frac{\theta_1}{2})(1 - e^{i \theta_2}) \\
    \cos(\frac{\theta_1}{2})(1 + e^{i \theta_2})
\end{matrix} \right] =
 \left[\begin{matrix}
    \sin(\frac{\theta_1}{2}) \sin(\theta_2) + i \sin(\frac{\theta_1}{2})(\cos(\theta_2) - 1) \\
    \sin(\frac{\theta_1}{2}) \sin(\theta_2) + i \sin(\frac{\theta_1}{2})(\cos(\theta_2) + 1) \\
    \cos(\frac{\theta_1}{2})(1 - \cos(\theta_2)) - i \cos(\frac{\theta_1}{2}) \sin(\theta_2)  \\
    \cos(\frac{\theta_1}{2})(1 + \cos(\theta_2)) + i \cos(\frac{\theta_1}{2}) \sin(\theta_2)  \\
\end{matrix} \right] 
\end{split}
\end{equation}
, where 
\begin{equation}
\ket{1, 1} = 
    \left[\begin{matrix}
   0 & 0 & 0 & 1 \\
\end{matrix} \right]^{T}, \nonumber
\end{equation}
and
\begin{equation}
\begin{split}
\label{eq:app_f_neg}
   f_{\bm{\theta}} (\tilde{\bm{x}}) &= \bra{1, 1} U^{\dagger}_{\bm{\theta}} | Z \otimes I |U_{\bm{\theta}} \ket{1, 1}\\ &= \sin^2(\frac{\theta_1}{2}) - \cos^2(\frac{\theta_1}{2}). 
\end{split}
\end{equation}
So $f_{\bm{\theta}} (\bm{x}) = f_{\bm{\theta}} (\tilde{\bm{x}})$. \hfill $\square$

\subsection{Sketch of Proof for Theorem 2}
\label{sec:app_thm2}
The proof is similar to \ref{sec:app_thm1}. Again, we prove the fundamental case where there is one readout qubit and one data qubit for QNN (\textit{XX-ZZ}) with Z-measurement. 

Following Eq.~5, we have
\begin{equation}
g_{\bm{\theta}} (\bm{x}) = \bra{1, 0} U^{\dagger}_{\bm{\theta}} | Z \otimes Z |U_{\bm{\theta}} \ket{1, 0},
\label{eq:app_g}
\end{equation}
where
\begin{equation}
Z \otimes Z = 
\left[\begin{matrix}
   1 & 0 & 0 & 0 \\
   0 & -1 & 0 & 0 \\
   0 & 0 & 1 & 0 \\
   0 & 0 & 0 & -1 \\
\end{matrix} \right]. \nonumber
\end{equation}
Substitute $Z \otimes Z$ and $U_{\bm{\theta}} \ket{1, 0}$ into Eq.~\ref{eq:app_g}, we have
\begin{equation}
g_{\bm{\theta}} (\bm{x}) = (\sin^2(\frac{\theta_1}{2}) - \cos^2(\frac{\theta_1}{2})) \cos(\theta_2)
\end{equation}.
Similarly, we have 
\begin{equation}
\begin{split}
g_{\bm{\theta}} (\tilde{\bm{x}}) &= \bra{1, 1} U^{\dagger}_{\bm{\theta}} | Z \otimes Z |U_{\bm{\theta}} \ket{1, 1} \\&= - (\sin^2(\frac{\theta_1}{2}) - \cos^2(\frac{\theta_1}{2})) \cos(\theta_2).
\end{split}
\end{equation}
So $g_{\bm{\theta}} (\bm{x}) = - g_{\bm{\theta}} (\tilde{\bm{x}})$. \hfill $\square$

\section{Classical Neural Networks}
\label{sec:app_exp}
For a comprehensive understanding of QNNs, we will show that negational symmetry is a unique property of QNNs when considering the large family of classical NNs. Here, we compare QNNs with 2 categories of classical models. Note, we choose simple models to validate negational symmetry instead of showing high performances. The first category is Deep Neural Networks (DNNs). The 2-layer DNN $D_1$ (\textit{16 - 2 - 1}) in Section~\ref{sec:exp_mnist} is treated as the baseline. We investigate 3 variants of the baseline: (i) increasing the number of nodes in the hidden layer $D_2$ (\textit{16 - 16 - 1}); (ii) increasing the number of hidden layers $D_3$ (\textit{16 - 16 - 2 - 1}); and (iii) increasing both the number of nodes in the hidden layer and the number of hidden layers $D_4$(\textit{16 - 16 - 16 - 1}). 
The second category is Convolutional Neural Networks (CNNs), which is a strong baseline. Considering the image resolution is only $4 \times 4$, we only use CNNs with simple architectures. The convolution operation is padded to have the same input and output feature map size. Each convolutional layer is followed by a ReLU activation function. We use max pooling to downscale the image size and use global average pooling to extract features from each feature channel. For a fair comparison and to avoid overfitting, we set the number of feature channels for a convolutional layer to be~16. We consider a 2-layer CNN $C_1$ and a 3-layer CNN $C_2$.
Let \textit{CV} stands for 1D convolutional layer with filter size $3$ and stride $1$, and \textit{FC} stand for fully-connected layer. Given an example, the input for CNNs is a 16-element binary vector and the output is a logit. The architecture of $C_1$ is \textit{CV-FC} and the architecture of $C_2$ is \textit{CV-CV-FC}. 
\end{document}